# Value of Information Lattice: Exploiting Probabilistic Independence for Effective Feature Subset Acquisition

**Mustafa Bilgic**                                                                                          mbilgic@iit.edu
*Illinois Institute of Technology*
*Chicago, IL 60616 USA*

**Lise Getoor**                                                                                            getoor@cs.umd.edu
*University of Maryland*
*College Park, MD 20742 USA*

## Abstract

We address the cost-sensitive feature acquisition problem, where misclassifying an instance is costly but the expected misclassification cost can be reduced by acquiring the values of the missing features. Because acquiring the features is costly as well, the objective is to acquire the right set of features so that the sum of the feature acquisition cost and misclassification cost is minimized. We describe the Value of Information Lattice (`VOILA`), an optimal and efficient feature subset acquisition framework. Unlike the common practice, which is to acquire features greedily, `VOILA` can reason with subsets of features. `VOILA` efficiently searches the space of possible feature subsets by discovering and exploiting conditional independence properties between the features and it reuses probabilistic inference computations to further speed up the process. Through empirical evaluation on five medical datasets, we show that the greedy strategy is often reluctant to acquire features, as it cannot forecast the benefit of acquiring multiple features in combination.

## 1. Introduction

We often need to make decisions and take appropriate actions in a complex and uncertain world. An important subset of decisions can be formulated as a *classification* problem, where an instance is described by a set of features and one of finite categorical options is chosen based on these features. Examples include medical diagnosis where the patients are described by lab tests and a diagnosis is to be made about the disease state of the patient, and spam detection where an email is described by its content and the email client needs to decide whether or not the email is spam.

Much research has been done on how to learn effective and efficient classifiers assuming that the features describing the entities are fully given. Even though this complete data assumption might hold on a few domains, in practice features that describe the entities often have missing values. In certain domains such as medical diagnosis where a decision is made based on a number of features that include laboratory test results, the missing feature values can be acquired at a cost by performing the related tests. In such cases, we need to decide which tests to perform in which order. The answer to this question, of course, depends on how important it is to get the correct classification decision. Put alternatively, the cost of an incorrect classification (e.g., a misdiagnosis) determines how much we are willing to spend on expensive tests. Thus, we need to devise a feature acquisition policy that can determine which tests to perform in which order and when to stop and make the final classification decision so that the total incurred cost, the feature acquisition cost and the expected misclassification cost, is minimized.

Devising the optimal policy in general requires considering all possible permutations of the features and their expected values. To provide some intuition, some features might be useful only if acquired together, and the cost and benefit of acquiring some features can depend on which other features have been acquired and what their values turned out to be. Because devising the optimal policy is intractable in general, previous work has been greedy (Gaag & Wessels, 1993; Yang, Ling, Chai, & Pan, 2006), has approximated value of information calculations (Heckerman, Horvitz, & Middleton, 1993), and has developed heuristic feature scoring techniques (Núñez, 1991; Turney, 1995).

The greedy approach, however, has at least two major limitations. First, because it considers each feature in isolation, it cannot accurately forecast the value of acquiring multiple features together, causing it to produce sub-optimal policies. Second, the greedy strategy assumes that features can be acquired sequentially and the value of a feature can be observed before acquiring the next one. This assumption, however, is often not very practical. For example, doctors typically order batches of measurements simultaneously such as blood count, cholesterol level, etc., and then possibly order another batch after the results arrive. These two limitations of the greedy approach make it necessary to reason with sets of features.

Reasoning with sets of features, on the other hand, poses serious tractability challenges. First of all, the number of subsets is exponential in the size of the feature set. Second, judging the value of acquiring a set of features requires taking an expectation over the possible values of the features in the set, which is also exponential in the number of the features. The good news, however, is that we do not need to consider all possible subsets of features in practice; certain features can render other features useless, while some features are useful only if acquired together. For example, an X-Ray might render a skin test useless for diagnosing tuberculosis. Similarly, a chest pain alone might not be useful for differentiating between a cold and a heart disease; it becomes useful only if it is combined with other features, such as a blood test.

In this article, we describe a data structure that discovers and exploits these types of constraints (features that render other features useless and features that are useful only if acquired together) from the underlying probability distribution. We propose *Value of Information Lattice* (`VOILA`) that reduces the space of all possible subsets by exploiting the constraints between the features. Additionally, `VOILA` makes it possible to share value of information calculations between different feature sets to reduce the computation time.

This article builds upon our earlier work (Bilgic & Getoor, 2007). Our contributions in this article include:

- We introduce two additional techniques for sharing computations between different subsets of features. These new techniques are based on information caching and utilizing paths in the underlying Bayesian network.

- We experiment with asymmetric misclassification costs in addition to the symmetric costs. The asymmetric setup reflects a more realistic case and provides new insights.

- In addition to the feature acquisition costs defined by Turney (1995), we generate and experiment with synthetic feature costs. The synthetic feature costs capture more complex feature acquisition costs and allows for leeway for various acquisition strategies to differ.

The remainder of this article is organized as follows. We describe the notation and problem formulation in Section 2. We describe how we can reduce the search space and share computations using `VOILA` in Section 3. We show experimental results in Section 4, discuss related work in Section 5, and discuss future work in Section 6. We then conclude in Section 7.

## 2. Notation and Problem Formulation

Our main task is to classify a given instance that has missing feature values and incur the minimum acquisition and misclassification cost. Let the instance be described by a set of features $\mathbf{X} = \{X_1, X_2, \ldots, X_n\}$ and let $Y$ be the random variable representing its class. We assume that the joint probability distribution $P(Y, \mathbf{X})$ is given and concern ourselves with feature acquisition during only inference (note that the conditional distribution $P(Y|\mathbf{X})$ is not appropriate, as most features are assumed to be unobserved initially). For the purpose of this article, we assume that we are given a Bayesian network, but any joint probabilistic model that allows us to efficiently answer conditional independence queries can be used.

In the notation, a bold face letter represents a set of random variables and non-bold face letter represents a single random variable. For example $\mathbf{X}$ represents the set of features, whereas $X_i \in \mathbf{X}$ represents a feature in $\mathbf{X}$ and $Y$ represents the class variable. Additionally, a capital letter represents a random variable, and a lowercase letter represents a particular value of that variable; this applies to both individual variables and sets of variables. For example, $Y$ represents a variable, and $y$ represents a particular value that $Y$ can take.

In addition to the probabilistic model, we are also given the cost models that specify feature acquisition costs and misclassification costs. Formally, we assume that we have a feature acquisition cost function that given a subset of features, $\mathbf{S}$, and the set of features whose values are known (evidence) $\mathbf{E}$, returns a non-negative real number $C(\mathbf{S} \mid \mathbf{e})$. We also assume that we have a misclassification cost model that returns the misclassification cost $c_{ij}$ incurred when $Y$ is assigned $y_i$ when the correct assignment is $y_j$. With these cost functions, we can model non-static feature acquisition costs; that is, the cost of acquiring the feature $X_i$ can depend on what has been acquired so far and what their values are ($\mathbf{e}$) as well what is acquired in conjunction with this feature ($\mathbf{S} \setminus \{X_i\}$). Moreover, the misclassification cost model does not assume symmetric costs; different kids of errors (false positives or negatives) can have different costs.

Figure 1 shows a simple example configuration with two features, $X_1$ and $X_2$, and the class variable $Y$. In this simple example, the joint distribution $P(\mathbf{X}, Y)$ is represented as a table, the feature costs are simple independent costs for $X_1$ and $X_2$, and the misclassification cost is symmetric where both types of misclassifications cost the same and correct classification does not cost anything.

A diagnostic policy $\pi$ is a decision tree where each node represents a feature and the branches from the nodes represent the possible values of the features. Each path of the policy, $\mathbf{p_s} \in \pi$, represents an *ordered* sequence feature values $\mathbf{s}$. We will often use $\mathbf{p_s}$ to represent an ordered version of $\mathbf{s}$. Typically, the order of the features in the set will be important for computing the feature costs, as the cost of a feature can depend on the values of previously acquired features. The order of the features will be irrelevant for computing the probability $P(\mathbf{s})$. An example conditional policy using the example configuration of Figure 1 is given in Figure 2.

| $X_1$ | $X_2$ | $Y$ | $P(X_1,X_2,Y)$ |
|---|---|---|---|
| T | T | T | 0.144 |
| T | T | F | 0.036 |
| T | F | T | 0.168 |
| T | F | F | 0.252 |
| F | T | T | 0.020 |
| F | T | F | 0.180 |
| F | F | T | 0.020 |
| F | F | F | 0.180 |

| Features | Cost |
|---|---|
| $X_1$ | |
| $X_1 \mid X_2 =T$ | 5 |
| $X_1 \mid X_2 =F$ | |
| $X_2$ | |
| $X_2 \mid X_1 =T$ | 10 |
| $X_2 \mid X_1 =F$ | |
| $X_1, X_2$ | 15 |

| Misclassification Costs | | Actual Label | |
|---|---|---|---|
|  |  | T | F |
| Predicted Label | T | 0 | 50 |
| | F | 50 | 0 |

Figure 1: Example configuration with two features $X_1$ and $X_2$ and class variable $Y$. The table from left to right represent: the joint probability distribution $P(X_1, X_2, Y)$, the feature costs, and the misclassification costs.

Each policy $\pi$ has two types of costs: the feature acquisition cost and a misclassification cost. These costs are defined in terms of the costs associated with following the paths in the policy. We first describe how to compute the feature acquisition cost of a path and then describe how to compute the associated expected misclassification cost. Finally, we show how to compute the expected total cost of a policy using the total costs associated with each path.

In the most naive version, the feature cost of a path $\mathbf{p_s}$ is the sum of the costs of the features that appear on the path. However, in practice, the cost of a feature can depend on which features have been acquired so far and the observed values of the acquired features. For example, performing the treadmill test (asking the patient to run a treadmill and measure his heart beat, etc.) can be riskier if we had ordered a cholesterol test and its result turned out to be high, putting the patient in high risk for heart disease. To account for these types of costs, the order of the features in $\mathbf{p_s}$ matters, and the total feature cost of a path is the summation of the individual feature costs conditioned on the values of the features that precede the features in consideration:

$$FC(\mathbf{p_s}) = \sum_{j=1}^{n} C(\mathbf{p_s}[j] \mid \mathbf{p_s}[1:j])$$

where $\mathbf{p_s}[j]$ represents the $j^{th}$ feature in $\mathbf{p_s}$ and $\mathbf{p_s}[1:j]$ represents feature values 1 through $j$ in $\mathbf{p_s}$.

When we reach the end of a path, we need to make a classification decision. In this case, we simply utilize the Bayesian decision theory and choose the decision with minimum risk (i.e., misclassification cost). We find such a decision by using the probabilistic model to

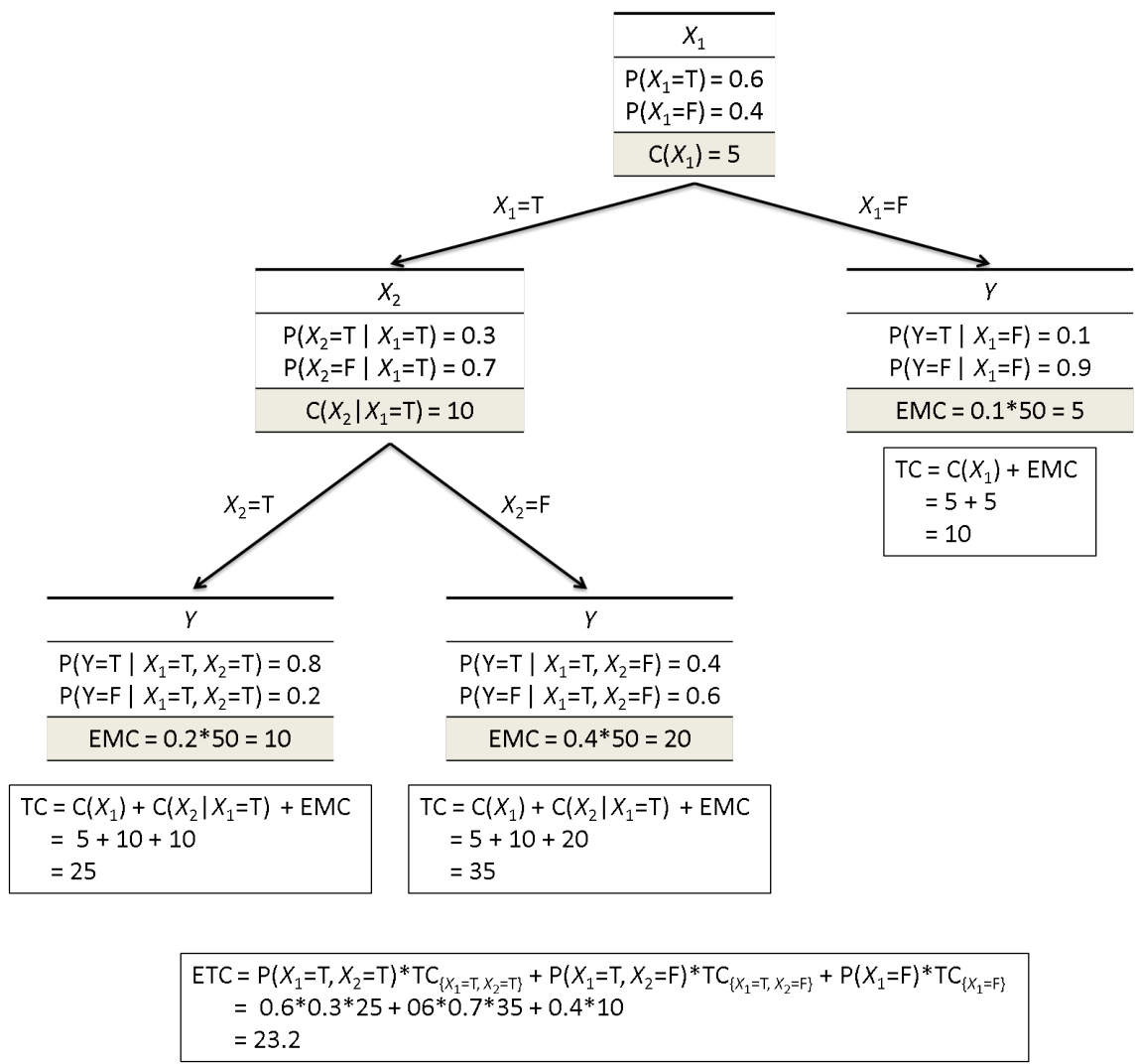

Figure 2: An example conditional policy with features $X_1, X_2$ and class variable $Y$. Each non-leaf node represents a feature acquisition, with probability distribution of the possible values, and the cost of the feature. Each path (e.g., $X_1 = T, X_2 = T$) has an acquisition cost and expected misclassification cost. The policy overall has an expected total cost $ETC$, which is the sum of total costs of each path, weighted by the probability of following that path.

compute the probability distribution $P(Y \mid \mathbf{p_s})$ and choose the value of $Y$ that leads to the minimum expected cost. Note that the order of the features values do not matter in this case; that is $P(Y \mid \mathbf{p_s}) = P(Y \mid \mathbf{s})$. The expected misclassification of a path, $EMC(\mathbf{p_s})$, is

computed as follows:

$$EMC(\mathbf{p_s}) = EMC(\mathbf{s}) = \min_{y_i} \sum_{y_j} P(Y = y_j \mid \mathbf{s}) \times c_{ij} \quad (1)$$

The total cost that we incur by following a path of a policy is simply the sum of the feature and the expected misclassification costs of that path:

$$TC(\mathbf{p_s}) = FC(\mathbf{p_s}) + EMC(\mathbf{p_s})$$

Finally, we compute the expected total cost of a policy $\pi$ using the total costs of the individual paths $\mathbf{p_s} \in \pi$. Each path $\mathbf{p_s} \in \pi$ has a probability of occurrence in real world. Such probability can be easily computed by the generative probability model that we assumed. It is simply $P(\mathbf{s})$. The expected total cost of a policy is then the sum of the total cost of each path, $TC(\mathbf{p_s})$, weighted by the probability of following that path, $P(\mathbf{s})$:

$$ETC(\pi) = \sum_{\mathbf{p_s} \in \pi} P(\mathbf{s}) TC(\mathbf{p_s}) \quad (2)$$

The objective of feature acquisition during inference is, given the joint probabilistic model and the cost models for acquisition and misclassification, find the policy that has the minimum expected total cost. However, building the optimal decision tree is known to be NP-complete (Hyafil & Rivest, 1976). Thus, most research have been greedy choosing the best feature that reduces the misclassification costs the most and has the lowest cost (e.g., Gaag & Wessels, 1993; Dittmer & Jensen, 1997) or have developed heuristic feature scoring techniques (e.g., Núñez, 1991; Tan, 1990).

In the greedy strategy, each path of the policy is extended with the feature that reduces the misclassification cost the most and that has the lowest cost. More specifically, the path $\mathbf{p_s}$ is replaced with new paths $\mathbf{p_{s \cup x_i^1}}, \mathbf{p_{s \cup x_i^2}}, \ldots, \mathbf{p_{s \cup x_i^n}}$ where $x_i^1, x_i^2, \ldots, x_i^n$ are the values that $X_i$ can take and $X_i$ is the feature that has the highest *benefit*. We define the *benefit* of a feature $X_i$ given a path $\mathbf{p_s}$ as the reduction in the total cost of the path when the path is expanded with the possible values of $X_i$. More formally,

$$Benefit(X_i \mid \mathbf{p_s}) \triangleq TC(\mathbf{p_s}) - \sum_{j=1}^{n} P(x_i^j \mid \mathbf{s}) TC(\mathbf{p_{s \cup x_i^j}})$$

$$= FC(\mathbf{p_s}) + EMC(\mathbf{s}) - \sum_{j=1}^{n} P(x_i^j \mid \mathbf{s}) \left( FC(\mathbf{p_{s \cup x_i^j}}) + EMC(\mathbf{s} \cup x_i^j) \right)$$

$$= FC(\mathbf{p_s}) - \left( \sum_{j=1}^{n} P(x_i^j \mid \mathbf{s}) FC(\mathbf{p_{s \cup x_i^j}}) \right) + EMC(\mathbf{s}) - \sum_{j=1}^{n} P(x_i^j \mid \mathbf{s}) EMC(\mathbf{s} \cup x_i^j)$$

$$= FC(\mathbf{p_s}) - (FC(\mathbf{p_s}) + C(X_i \mid \mathbf{s})) + EMC(\mathbf{s}) - \sum_{j=1}^{n} P(x_i^j \mid \mathbf{s}) EMC(\mathbf{s} \cup x_i^j)$$

$$= -C(X_i \mid \mathbf{s}) + EMC(\mathbf{s}) - \sum_{j=1}^{n} P(x_i^j \mid \mathbf{s}) EMC(\mathbf{s} \cup x_i^j)$$

Note that, the last two terms are equivalent to the definition of expected value of information, *EVI*, (Howard, 1966):

$$EVI(X_i \mid \mathbf{s}) = EMC(\mathbf{s}) - \sum_{j=1}^{n} P(x_i^j \mid \mathbf{s}) EMC(\mathbf{s} \cup x_i^j) \qquad (3)$$

Substituting *EVI*, the definition of benefit becomes very intuitive:

$$Benefit(X_i \mid \mathbf{p_s}) = Benefit(X_i \mid \mathbf{s}) = EVI(X_i \mid \mathbf{s}) - C(X_i \mid \mathbf{s}) \qquad (4)$$

With this definition, the greedy strategy iteratively finds the feature that has the highest positive benefit (value cost difference), acquires it, and stops acquisition when there are no more features with a positive benefit value.

We also note that it is straightforward to define *EVI* and *Benefit* for a set $\mathbf{S}'$ of features just like we did for a single feature. The only difference is that the expectation needs to be taken over the joint assignments, $\mathbf{s}'$, to the features in the set $\mathbf{S}'$.

$$EVI(\mathbf{S}' \mid \mathbf{s}) = EMC(\mathbf{s}) - \sum_{\mathbf{s}'} P(\mathbf{s}' \mid \mathbf{s}) EMC(\mathbf{s} \cup \mathbf{s}') \qquad (5)$$

and,

$$Benefit(\mathbf{S}' \mid \mathbf{s}) = EVI(\mathbf{S}' \mid \mathbf{s}) - C(\mathbf{S}' \mid \mathbf{s}) \qquad (6)$$

There are a few problems with the greedy strategy as we have mentioned earlier. First, it is short-sighted. There exist sets $\mathbf{S} \subseteq \mathbf{X}$ such that $Benefit(\mathbf{S}) > \sum_{X_i \in \mathbf{S}} Benefit(X_i)$. This is easier to see, for example, for the XOR function, $Y = X_1$ XOR $X_2$, where $X_1$ and $X_2$ alone are not useful but they are determinative together. Due to this relationship, a greedy policy is not guaranteed to be optimal. Moreover, the greedy policy can prematurely stop acquisition because no single feature seems to provide positive benefit.

The second problem with the greedy strategy is that we often need to acquire a set of features simultaneously. For example, a doctor orders a set of lab tests when s/he sends the patient to a lab, such as blood count, cholesterol level, etc. rather than ordering a single test, waiting for its result and ordering the next one. However, the traditional greedy strategy cannot handle reasoning with sets of features naturally.

We would like to be able to reason with sets of features for these two reasons. Our objective in this article is, given an existing potentially empty set of already observed features $\mathbf{E}$ and their observed values $\mathbf{e}$, find the set that has the highest benefit:

$$\mathbf{L}(\mathbf{X} \mid \mathbf{e}) \triangleq \operatorname*{argmax}_{\mathbf{S} \subseteq \mathbf{X} \setminus \mathbf{E}} Benefit(\mathbf{S} \mid \mathbf{e}) \qquad (7)$$

There are two problems with this formulation: first, the number of subsets of $\mathbf{X} \setminus \mathbf{E}$ is exponential in the size of $\mathbf{X} \setminus \mathbf{E}$, and second, for each set $\mathbf{S}$, we need to take an expectation over the joint assignments to all features in the set. We address these two problems using a data structure that we describe next.

## 3. Value of Information Lattice (VOILA)

VOILA makes reasoning with sets of features tractable by reducing the space of possible sets and allowing sharing of *EVI* computations between different sets. In this section, we will first explain how we can reduce the space and then explain techniques for computation sharing.

### 3.1 Reducing the Space of Possible Sets

In most domains, there are often complex interactions between the features and the class label. Contrary to the Naive Bayes assumption, features are often not conditionally independent given the class label. Some features are useless once some other features are already acquired. For example a chest X-Ray is typically more determinative than a skin test for tuberculosis. Similarly, some features are useless alone unless they are accompanied with other features. For example, a chest pain alone might be due to a variety of sicknesses; if it is accompanied with high cholesterol, it could indicate a heart disease, whereas if it is combined with fever, a cold might be more probable. These types of interactions between the features allow us to reduce the space of candidate feature sets.

As we have mentioned in the problem formulation, we have assumed that we already have a joint probabilistic model over the features and the class variable, $P(Y, \mathbf{X})$. We will find these two types of feature interactions by asking probabilistic independence queries using $P(Y, \mathbf{X})$. Specifically, we assume that we are given a Bayesian network that represents $P(Y, \mathbf{X})$. The Bayesian network will allow us to find these types of interactions through standard d-separation algorithms.

**Definition 1** *A set $\mathbf{S} \subseteq \mathbf{X} \setminus \mathbf{E}$ is* **irreducible** *with respect to evidence $\mathbf{e}$ if $\forall X_i \in \mathbf{S}$, $X_i$ is <u>not</u> conditionally independent of $Y$ given $\mathbf{e}$ and $\mathbf{S} \setminus \{X_i\}$.*

Given a Bayesian network over $\mathbf{X}$ and $Y$, it is straightforward to check irreducibility through d-separation (Pearl, 1988).

**Proposition 1** *Let $\mathbf{S}'$ be a maximal irreducible subset of $\mathbf{S}$ with respect to $\mathbf{e}$. Then, $EVI(\mathbf{S} \mid \mathbf{e}) = EVI(\mathbf{S}' \mid \mathbf{e})$.*

*Proof:* Let $\mathbf{S}'' = \mathbf{S} \setminus \mathbf{S}'$. If $\mathbf{S}'$ is a maximal irreducible set, $\mathbf{S}' \cup \mathbf{E}$ d-separates $Y$ and $\mathbf{S}''$. Otherwise, we could make $\mathbf{S}'$ larger by including the non-d-separated element(s) from $\mathbf{S}''$ in $\mathbf{S}'$. Thus, we have $P(Y \mid \mathbf{e}, \mathbf{s}) = P(Y \mid \mathbf{e}, \mathbf{S}', \mathbf{S}'') = P(Y \mid \mathbf{e}, \mathbf{S}')$. Substitution in Equations 1 and 5 yields the desired property.

Note that under the assumption that $C(\mathbf{S}' \mid \mathbf{e}) \leq C(\mathbf{S} \mid \mathbf{e})$ for any $\mathbf{S}' \subseteq \mathbf{S}$, it suffices to consider only the irreducible sets to find the optimal solution to the objective function in Equation (7). VOILA is a data structure that contains only the irreducible feature subsets of $\mathbf{X}$, with respect to a particular set of evidence $\mathbf{e}$. We next define VOILA formally.

**Definition 2** *A* VOILA *$\mathbf{V}$ is a directed acyclic graph in which there is a node corresponding to each possible irreducible set of features, and there is a directed edge from a feature set $\mathbf{S}$ to each node that corresponds to a direct (maximal) subset of $\mathbf{S}$. Other subset relationships in the lattice are then defined through the directed paths in $\mathbf{V}$.*

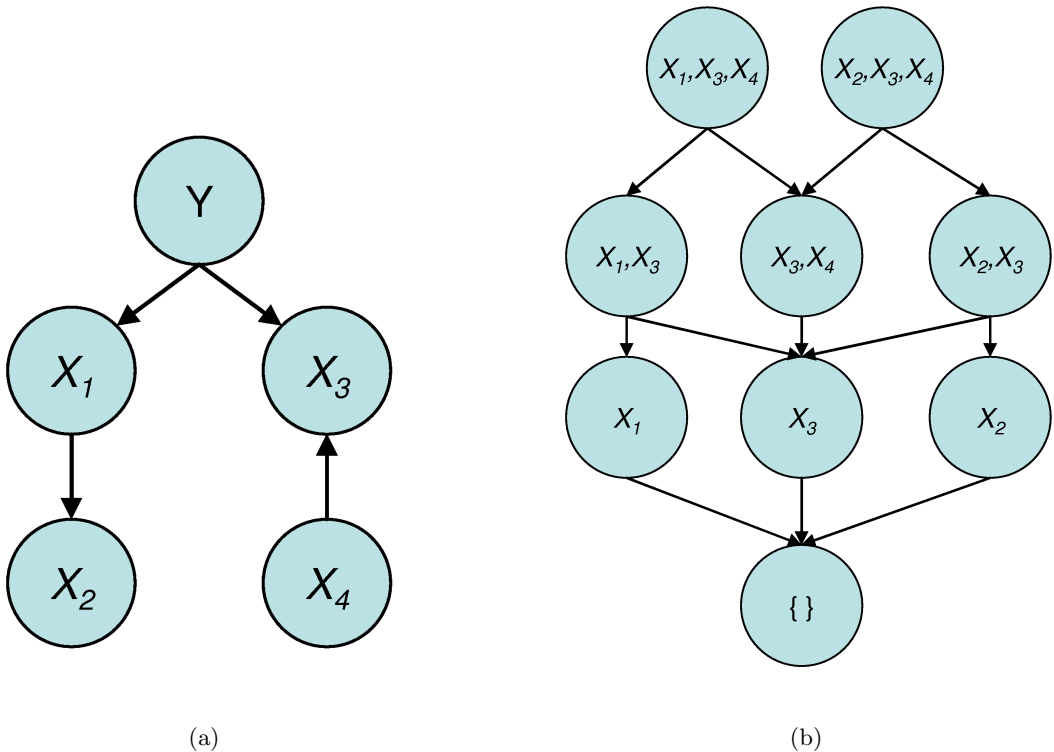

Figure 3: (a) A simple Bayesian network illustrating dependencies between attributes and the class variable. (b) The VOILA corresponding to the network.

Figure 3(a) shows a simple Bayesian network and its corresponding VOILA, with respect to the empty evidence set, is shown in Figure 3(b). Notice that the VOILA contains only the irreducible subsets given the Bayesian network; for instance, the VOILA does not contain sets that include both $X_1$ and $X_2$ because $X_1$ d-separates $X_2$ from $Y$. We also observe that the number of irreducible subsets is 9 in contrast to $2^4 = 16$ possible subsets. Moreover, note that the largest subset size is now 3 in contrast to 4. Having smaller feature sets sizes has a dramatic effect on the value of information calculations. In fact, these savings can make solving the objective function optimally (Equation (7)) feasible in practice.

### 3.2 Sharing *EVI* Calculations

Finding the set **S** that has the highest *Benefit* (Equation 6) requires computing $EVI(\mathbf{S})$ (Equation 5). However, computing $EVI(\mathbf{S})$ requires taking an expectation over all possible values of the features in **S**. Moreover, searching for the best set among all the irreducible sets requires us to compute $EVI$ for all irreducible sets. To make such computations tractable in practice, VOILA allows computation sharing between its nodes. In this article, we describe three possible ways of sharing computations between the nodes of VOILA.

### 3.2.1 Subset Relationships

`VOILA` exploits the subset relationships between different feature sets in order to avoid computing *EVI* for some nodes. First of all, if there is a directed path from node $\mathbf{S}_1$ to $\mathbf{S}_2$ in `VOILA`, then $S_1 \supset S_2$ and thus $EVI(\mathbf{S}_1 \mid \mathbf{e}) \geq EVI(\mathbf{S}_2 \mid \mathbf{e})$[1]. Now assume that there is a directed path from $\mathbf{S}_i$ to $\mathbf{S}_j$ and $EVI(\mathbf{S}_i \mid \mathbf{e}) = EVI(\mathbf{S}_j \mid \mathbf{e})$. Then, all of the nodes on this path will also have the same *EVI*, thus we do not need to do the computation for those subsets. An algorithm that makes use of this observation is given in Algorithm 1.

---

**Algorithm 1**: Efficient *EVI* computation using `VOILA`.

    **Input**: `VOILA` **V** and current evidence **E**
    **Output**: `VOILA` updated with correct *EVI* values
**1**  **for** *all root node(s)* **S**
**2**     $value \leftarrow EVI(\mathbf{S} \mid \mathbf{e}); ub(\mathbf{S}) \leftarrow value; lb(\mathbf{S}) \leftarrow value$
**3**     $ub(descendants(\mathbf{S})) \leftarrow value$
**4**  **for** *all leaf node(s)* **S**
**5**     $value \leftarrow EVI(\mathbf{S} \mid \mathbf{e}); ub(\mathbf{S}) \leftarrow value; lb(\mathbf{S}) \leftarrow value$
**6**     $lb(ancestors(\mathbf{S})) \leftarrow value$
**7**  **for** *all node* **S** *where* $lb(\mathbf{S}) \neq ub(\mathbf{S})$
**8**     $value \leftarrow EVI(\mathbf{S} \mid \mathbf{e}); ub(\mathbf{S}) \leftarrow value; lb(\mathbf{S}) \leftarrow value$
**9**     $lb(ancestors(\mathbf{S})) \leftarrow value$
**10**    $ub(descendants(\mathbf{S})) \leftarrow value$

---

It is important to point out that all nodes of `VOILA` are irreducible sets. Unless there are totally useless features that do not change $P(Y)$ when observed, then we should not have any two distinct nodes where the *EVI* values are exactly equal. However, this statement is true only if we do not have any context-specific independencies (independencies that hold only under certain assignments to the variables) in the underlying Bayesian network. In our description and implementation, we used standard d-separation at the variable level; one can imagine going one step further and define the irreducible sets through both the variable level d-separation and context specific independencies.

In order to share computations between different nodes of the lattice, we keep lower and upper bounds on the *EVI* of a node. The lower bound is determined by the values of the descendants of the node whereas the upper bound is determined by the values of its ancestors. First, we initialize these bounds by computing the value of the information at the boundary of the lattice, i.e., the root node(s) and the leaf node(s) (lines 1–6) [2]. Then, we loop over the nodes whose upper bounds and lower bounds are not equal (line 7–10), computing their values and updating the bounds at their ancestors and descendants. The algorithm terminates when the upper bounds and lower bounds for all the nodes become tight. The order in which to choose the nodes in line 7 so that the number of sets for which a value is calculated is minimum is still an open question. A possible heuristic is to perform a binary search and choose a middle node on a path between two nodes for which the values have already been calculated.

---

1. A superset has always a higher or equivalent *EVI* (Equation (5)) than its subset.
2. We do not need to compute *EVI* for all root nodes; it suffices to compute it for the node that corresponds to the Markov blanket of $Y$. This will be explained in more detail in the next section.

### 3.2.2 INFORMATION PATHWAYS AT THE UNDERLYING BAYESIAN NETWORK

The second mechanism that `VOILA` uses to share *EVI* computations is through the edges in the underlying Bayesian network. We specifically make use of the following fact:

**Proposition 2** *For all $\mathbf{S}_1$ and $\mathbf{S}_2$, if $\mathbf{S}_1$ d-separates $Y$ from $\mathbf{S}_2$ with respect to $\mathbf{e}$, then $EVI(\mathbf{S}_1 \mid \mathbf{e}) \geq EVI(\mathbf{S}_2 \mid \mathbf{e})$.*

*Proof:* Consider $\mathbf{S}_{12} = \mathbf{S}_1 \cup \mathbf{S}_2$. Because of the subset relationship, we know that $EVI(\mathbf{S}_{12} \mid \mathbf{e}) \geq EVI(\mathbf{S}_1 \mid \mathbf{e})$ and $EVI(\mathbf{S}_{12} \mid \mathbf{e}) \geq EVI(\mathbf{S}_2 \mid \mathbf{e})$.

$$\begin{aligned}
EVI(\mathbf{S}_{12} \mid \mathbf{e}) &= EMC(Y \mid \mathbf{e}) - \sum_{\mathbf{s}_{12}} P(\mathbf{s}_{12} \mid \mathbf{e}) EMC(Y \mid \mathbf{e}, \mathbf{s}_{12}) \\
&= EMC(Y \mid \mathbf{e}) - \sum_{\mathbf{s}_1}\sum_{\mathbf{s}_2} P(\mathbf{s}_1, \mathbf{s}_2 \mid \mathbf{e}) EMC(Y \mid \mathbf{e}, \mathbf{s}_1, \mathbf{s}_2) \\
&= EMC(Y \mid \mathbf{e}) - \sum_{\mathbf{s}_1}\sum_{\mathbf{s}_2} P(\mathbf{s}_1, \mathbf{s}_2 \mid \mathbf{e}) EMC(Y \mid \mathbf{e}, \mathbf{s}_1) \\
&= EMC(Y \mid \mathbf{e}) - \sum_{\mathbf{s}_1} P(\mathbf{s}_1 \mid \mathbf{e}) EMC(Y \mid \mathbf{e}, \mathbf{s}_1) \\
&= EVI(\mathbf{S}_1 \mid \mathbf{e}) \\
&\geq EVI(\mathbf{S}_2 \mid \mathbf{e})
\end{aligned}$$

The third line follows from the second by the fact that $\mathbf{S}_1$ d-separates $Y$ from $\mathbf{S}_2$ and thus $P(Y \mid \mathbf{s}_1, \mathbf{s}_2) = P(Y \mid \mathbf{s}_1)$.

**Corollary:** The Markov blanket of $Y$, (i.e., $Y$'s parents, $Y$'s children, and $Y$'s children's other parents), is the set that has the highest *EVI* in our search space, as it d-separates all of the remaining variables from $Y$. Using this corollary, we do not need to compute the *EVI* for all root nodes in Algorithm 1; we can compute *EVI* for the root node that corresponds to the Markov blanket of $Y$ and it serves as the upper bound for the *EVI* of the remaining root nodes.

These relationships can very well be exploited like we exploited the subset relationships above. Instead of just using the subset relationships, we can use both subset and independence relationships. One simple way to make use of Algorithm 1 without modification is to add edges between any $\mathbf{S}_1$ and $\mathbf{S}_2$ where the independence property holds. An example $\mathbf{S}_1$ and $\mathbf{S}_2$ according to our toy network in Figure 3(a) would be $\mathbf{S}_1 = \{X_1\}$ and $\mathbf{S}_2 = \{X_2\}$. Thus, we can add a directed edge from $X_1$ to $X_2$ in our `VOILA` in Figure 3(b) and Algorithm 1 will work just fine.

### 3.2.3 INCREMENTAL INFERENCE

The third and the last mechanism that `VOILA` uses for computation sharing is through caching of probabilities at its nodes. For each candidate set $\mathbf{S} \in \mathbf{V}$, we need to compute $EVI(\mathbf{S} \mid \mathbf{e})$ which requires computing $P(\mathbf{S} \mid \mathbf{e})$ and $EMC(Y \mid \mathbf{S}, \mathbf{e})$. If we cache the conditional probabilities at each node of $\mathbf{V}$, then to compute $P(\mathbf{S} \mid \mathbf{e})$, we find one of its supersets $\mathbf{S}_i = \mathbf{S} \cup \{X_i\}$ and then compute $P(\mathbf{S} \mid \mathbf{e}) = \sum_{x_i} P(\mathbf{S}, X_i = x_i \mid \mathbf{e})$.

Computing $EMC(Y \mid \mathbf{S}, \mathbf{e})$ requires computing $P(Y \mid \mathbf{S}, \mathbf{e})$. To perform this computation efficiently, we cache the state of the junction tree at each node of the VOILA. Then, we find a subset, $\mathbf{S}_j$, such that $\mathbf{S} = \mathbf{S}_j \cup \{X_j\}$. We compute $P(Y \mid \mathbf{S}, \mathbf{e})$ by integrating the extra evidence to the junction tree at node $\mathbf{S}_j$ that is used to compute $P(Y \mid \mathbf{S}_j, \mathbf{e})$.

### 3.3 Constructing VOILA

Efficient construction of VOILA is not a straightforward task. The brute force approach would be to enumerate all possible subsets of $\mathbf{X} \setminus \mathbf{E}$ and for each subset check whether it is irreducible. However, this brute force approach is clearly impractical. Because the number of nodes in VOILA is expected to be much fewer than the number of possible subsets of $\mathbf{X} \setminus \mathbf{E}$, if we can be smart about which sets we consider for inclusion in $\mathbf{V}$, we can construct it more efficiently. That is, instead of generating all possible candidates and checking whether they are irreducible or not, we try to generate only irreducible sets. We first introduce the notion of a dependency constraint and then explain how we can use dependency constraints to efficiently construct VOILA.

**Definition 3** *A* dependency constraint *for a feature $X_i \in \mathbf{S}$ with respect to $\mathbf{S}$ and $\mathbf{E}$ is the constraint on $\mathbf{S} \cup \mathbf{E}$ that ensures a dependency between $X_i$ and $Y$ exists.*

For instance, in our running example, a dependency constraint for $X_2$ is $\neg X_1$; in other words, in order for $X_2$ to be relevant, $X_1$ should not be included in $\mathbf{S} \cup \mathbf{E}$. Similarly, the dependency constraint for $X_4$ is $X_3$, meaning that $X_3$ must be included in $\mathbf{S} \cup \mathbf{E}$. Specifically, a dependency constraint for a feature $X_i$ requires that all $X_j$ on the path from $Y$ to $X_i$ not to be included in $\mathbf{S} \cup \mathbf{E}$ if $X_j$ is not part of a v-structure; if $X_j$ is part of a v-structure, then either $X_j$ or one of its descendants must be included in $\mathbf{S} \cup \mathbf{E}$ (we refer to these latter constraints as *positivity constraints*). The algorithm that uses these ideas to compute the dependency constraints for each feature is given in Algorithm 2.

---

**Algorithm 2**: Dependency constraint computation for $X_i$.

    **Input**: $X_i, Y$
    **Output**: Dependency constraint for $X_i$, denoted $DC(X_i)$
**1** $DC(X_i) \leftarrow$ false
**2** **for** *each undirected path $p_j$ between $X_i$ and $Y$*
**3**      $DC_j(X_i) \leftarrow$ true
**4**      **for** *each $X_k$ on the path $p_j$*
**5**          **if** *$X_k$ does not a cause a v-structure* **then**
**6**              $DC_j(X_i) \leftarrow DC_j(X_i) \wedge \neg X_k$
**7**          **else**
**8**              $DC_j(X_i) \leftarrow DC_j(X_i) \wedge (X_k \vee Descendants(X_k))$
**9**      $DC(X_i) \leftarrow DC(X_i) \vee DC_j(X_i)$

---

These dependency constraints can be used to check whether a set is irreducible or potentially irreducible. Intuitively, a set is potentially irreducible if it is not irreducible but it is possible to make the set irreducible by adding more features into it. More formally,

**Definition 4** *A set* $\mathbf{S} \subseteq \mathbf{X} \setminus \mathbf{E}$ *is* **potentially irreducible** *with respect to evidence* $\mathbf{e}$ *if,* $\mathbf{S}$ *is not irreducible but there exists a non-empty set of features* $\mathbf{S}' \subseteq \mathbf{X} \setminus \{\mathbf{E} \cup \mathbf{S}\}$ *such that* $\mathbf{S} \cup \mathbf{S}'$ *is irreducible.*

Potential irreducibility is possible due to the non-monotonic nature of d-separation. That is, a feature that is d-separated from $Y$ can become dependent if we consider it in combination with other features. For example, in our running example, $\{X_4\}$ is not irreducible, as $X_4$ is d-separated from $Y$, whereas $\{X_3, X_4\}$ is irreducible.

We use the dependency constraints to check whether a set is irreducible or potentially irreducible. Because a set $\mathbf{S}$ is irreducible only if a dependency between all of its elements and $Y$ exists, the dependency constraint for the set $\mathbf{S}$ is the conjunction of the dependency constraints of its members. The irreducibility of $\mathbf{S}$ can be checked by setting the elements of $\mathbf{S}$ and $\mathbf{E}$ to "true" and setting the remaining elements of $\mathbf{X}$ to "false" and evaluating the set's dependency constraint. In our running example, the dependency constraint for the set $\{X_2, X_4\}$ is $\neg X_1 \wedge X_3$. Assuming $\mathbf{E} = \emptyset$, when we set the members of $\{X_2, X_4\}$ to true, and set the remaining features, $X_1$ and $X_3$, to false, $\neg X_1 \wedge X_3$ then evaluates to false and thus this set is not irreducible. This makes sense because given no evidence, $X_4$ is independent of $Y$, so while $\{X_2\}$ is a useful feature set to consider for acquisition, $\{X_2, X_4\}$ is not.

Checking for potential irreducibility is very similar. Set the elements of $\mathbf{S}$ and $\mathbf{E}$ to true like we did above. Then, set the positivity constraints of the members of $\mathbf{S}$ to true. Finally, set everything else to false. Using the same example above, to check whether $\{X_2, X_4\}$ is potentially irreducible, set $X_2 = $ true, $X_4 = $ true. Also set $X_3 = $ true because it is the positivity constraint for $X_4$. Set the remaining features, that is $X_1$, to false. Evaluating the constraint $\neg X_1 \wedge X_3$ yields to true, showing that $\{X_2, X_4\}$ is potentially irreducible (while it was not irreducible).

Given the definitions of irreducibility and potential irreducibility and the mechanisms to check for these properties through the notion of dependency constraints, we next describe the algorithm to construct VOILA.

VOILA construction proceeds in a bottom up fashion, beginning with the lowest level, which initially contains only the empty set and constructs new irreducible feature sets by adding one feature at a time into the VOILA structure. Algorithm 3 gives the details of the algorithm. The algorithm keeps track of the irreducible feature sets **IS**, and the set of potentially irreducible feature sets **PS**. When we are done processing feature $X_{i_j}$, we remove from **PS** any potentially irreducible set that cannot become irreducible because $X_{i_j}$ will not be re-considered (line 11).

### 3.3.1 ANALYSIS OF VOILA CONSTRUCTION ALGORITHM

The construction algorithm inserts a node into the VOILA only if the corresponding set is irreducible (lines 6 and 7). Moreover, by keeping track of potentially irreducible sets (lines 8–10), we generate every possible irreducible set that can be generated. Thus, VOILA contains only *and* all of the possible irreducible subsets of $\mathbf{X}$.

---

**Algorithm 3**: The VOILA construction algorithm.
    **Input**: Set of features $\mathbf{X}$ and class variable $Y$.
    **Output**: The VOILA data structure $\mathbf{V}$, given $\mathbf{E}$.
**1** Pick an ordering of elements of $\mathbf{X} = X_{i_1}, X_{i_2}, \ldots, X_{i_n}$
**2** $\mathbf{IS} \leftarrow \{\emptyset\}; \mathbf{PS} \leftarrow \emptyset$
**3** **for** $j = 1$ to $n$
**4**     **for** each $\mathbf{S} \in \mathbf{IS} \cup \mathbf{PS}$
**5**         $\mathbf{S}' \leftarrow \mathbf{S} \cup X_{i_j}; DC(\mathbf{S}') \leftarrow DC(\mathbf{S}) \wedge DC(X_{i_j})$
**6**         **if** $\mathbf{S}'$ *is irreducible* **then**
**7**             $\mathbf{IS} \leftarrow \mathbf{IS} \cup \{\mathbf{S}'\}$; Add a node corresponding to $\mathbf{S}'$ to $\mathbf{V}$
**8**         **else**
**9**             **if** $\mathbf{S}'$ *is potentially irreducible* **then**
**10**                 $\mathbf{PS} \leftarrow \mathbf{PS} \cup \{\mathbf{S}'\}$
**11**     Remove from $\mathbf{PS}$ all sets that are no longer potentially irreducible
**12** max = size of largest $\mathbf{S}$ in $\mathbf{IS}$; $L_l = \{S \mid S \in \mathbf{IS} \text{ and } |S| = l\}$
**13** **for** $l = 0$ to $max - 1$
**14**     **for** each $\mathbf{S} \in \mathbf{L}_l$
**15**         **for** each $\mathbf{S}' \in \mathbf{L}_{l+1}$
**16**             **if** $\mathbf{S} \subset \mathbf{S}'$ **then**
**17**                 Add an edge from $\mathbf{S}'$ to $\mathbf{S}$ to $\mathbf{V}$

---

The worst-case running time of the algorithm is still exponential in the number of initially unobserved features, $\mathbf{X} \setminus \mathbf{E}$, because number of irreducible sets can potentially be exponential. The running time in practice, though, depends on the structure of the Bayesian network that the VOILA is based upon and the ordering of the variables in line 1. For example, if the Bayesian network is naive Bayes, then all subsets are irreducible (no feature d-separates any other feature from the class variable); thus, the search space cannot be reduced at all. However, naive Bayes makes extremely strong assumptions which are unlikely to hold in practice. In fact, as we empirically show in the experiments section on five real-world datasets, features often are *not* conditionally independent given the class variable; there are more complex interactions between them and thus the number of irreducible subsets is substantially smaller than the number of all possible subsets.

The for loop at line 4 iterates over each irreducible and potentially irreducible sets that have been generated so far, and the number of potentially-irreducible sets generated depends on the ordering chosen. A good ordering processes features with literals with positivity constraints in other features' dependency constraints earlier. That is, for each undirected path from $Y$ to $X_i$ that includes $X_j$ in a v-structure, a good ordering puts $X_j$ earlier in the ordering than everything between $X_j$ and $X_i$. For instance, in our sample Bayesian network in Figure 3(a), we should consider $X_3$ earlier than $X_4$. We refer to an ordering as *perfect* if it satisfies all the positivity constraints. If a perfect ordering is used, VOILA construction algorithm never generates a potentially irreducible set. Unfortunately, it is not always possible to find a perfect ordering. A perfect ordering is not possible when two features have each other as a positivity constraint literal in their dependency constraints. This case occurs only when there is a loop from $Y$ to $Y$ that has two or more v-structures (Note that even though a Bayesian network is a directed acyclic graph, it can still contain loops, i.e., undirected cycles). A perfect ordering was possible in four of the five real world datasets that we used.

### 3.4 Using VOILA for Feature-value Acquisition

VOILA makes searching the space of all possible subsets tractable in practice. Using this flexibility, it is possible to devise several different acquisition policies. We describe two policies as example policies in this section.

The first acquisition policy aims to capture the practical setting where more than one feature is acquired at once. The policy can be constructed using VOILA as follows. Each path $\mathbf{p_s}$ of the policy $\pi$ (which is initially empty) is repeatedly extended by acquiring the set $\mathbf{S'} \in \mathbf{V}$ that has the best $Benefit(\mathbf{S'} \mid \mathbf{s}, \mathbf{e})$. The policy construction ends when no path can be extended, i.e., all candidate sets have non-positive $Benefit$ values for each path of $\pi$.

The second acquisition policy adds a look-ahead capability to the greedy policy. That is, rather than repeatedly extending each path $\mathbf{p_s}$ of policy $\pi$ with the feature $X_i$ that has the highest $Benefit(X_i \mid \mathbf{s}, \mathbf{e})$, we add a look-ahead capability, and first find the set $\mathbf{S'} \in \mathbf{V}$ that has the highest $Benefit(\mathbf{S'} \mid \mathbf{s}, \mathbf{e})$. Then, instead of acquiring all features in $\mathbf{S'}$ all at once, like we did in the above policy, we find the feature $X_i \in \mathbf{S'}$ that has the highest $Benefit(X_i \mid \mathbf{s}, \mathbf{e})$ and acquire it to extend $\mathbf{p_s}$.

## 4. Experiments

We experimented with five real-world medical datasets that Turney (1995) described and used in his paper. These datasets are Bupa Liver Disorders, Heart Disease, Hepatitis, Pima Indians Diabetes, and Thyroid Disease, which are all available from the UCI Machine Learning Repository (Frank & Asuncion, 2010). The datasets had a varying number of features ranging from five to 20. Four out of five datasets had binary labels, whereas the Thyroid dataset had three labels.

For each dataset, we first learned a Bayesian Network that both provides the joint probability distribution $P(Y, \mathbf{X})$ and efficiently answers conditional independence queries thorough d-separation (Pearl, 1988). We built a VOILA for each dataset using the learned Bayesian Network. We first present statistics on each dataset, such as the number of features and number of nodes in the VOILA, and then compare various acquisition policies.

### 4.1 Search Space Reduction

Table 1 shows aggregate statistics about each dataset, describing the number of features, the number of all possible subsets, the number of subsets in VOILA, and the percent reduction in the search space. As this table shows, the number of irreducible subsets is substantially fewer than all possible subsets. For the Thyroid Disease dataset, for example, the number of possible subsets is over a million whereas the number of irreducible subsets is fewer than thirty thousand. This enormous reduction in the search space makes searching through the possible sets of features tractable in practice.

Table 1: Aggregate statistics about each dataset. The number of irreducible subsets, i.e., the number of nodes in VOILA, is substantially fewer than the number of all possible subsets.

| Dataset | Features | All Subsets | Nodes in VOILA | Reduction |
| --- | --- | --- | --- | --- |
| Bupa Liver Disorders | 5 | 32 | 26 | 19% |
| Pima Indians Diabetes | 8 | 256 | 139 | 46% |
| Heart Disease | 13 | 8,192 | 990 | 88% |
| Hepatitis | 19 | 524,288 | 18,132 | 97% |
| Thyroid Disease | 20 | 1,048,576 | 28,806 | 97% |

### 4.2 Expected Total Cost Comparisons

We compared the expected total costs (Equation 2) of four different acquisition policies for each dataset. These policies are as follows:

- No Acquisition: This policy does not acquire any features; it aims to minimize the expected misclassification cost based on the prior probability distribution of the class variable, $P(Y)$.

- Markov Blanket: This policy acquires every relevant feature, regardless of the misclassification costs. The Market Blanket of $Y$ in a Bayesian network is defined as $Y$'s parents, children, and its children's other parents (Pearl, 1988). Intuitively, it is the minimal set $\mathbf{S} \subseteq \mathbf{X}$ such that $Y \perp (\mathbf{X} \setminus \mathbf{S}) \mid \mathbf{S}$.

- Greedy: This policy repeatedly expands each path $\mathbf{p_s}$ of an initially empty policy $\pi$ by acquiring the feature $X_i$ that has the highest positive $Benefit(X_i \mid \mathbf{s})$ (Equation 4). The policy construction ends when no path can be extended with a feature with a positive $Benefit$ value.

- Greedy-LA: This policy adds a look-ahead capability to the Greedy strategy. This policy repeatedly expands each path $\mathbf{p_s}$ of an initially empty policy $\pi$ by first finding the set $\mathbf{S}'$ that has the highest positive $Benefit(\mathbf{S}' \mid \mathbf{s})$ (Equation 6) and then acquiring the feature $X_i \in \mathbf{S}'$ that has the maximum $Benefit(X_i \mid \mathbf{s})$ (Equation 4). The policy construction ends when no set with a positive $Benefit$ value can be found for any path of the policy.

The feature costs for each dataset are described in detail by Turney (1995). In summary, each feature can either have an independent cost, or can belong to a group of features, where the first feature in that group incurs an additional cost. For example, the first feature from a group of blood measurements incurs the overhead cost of drawing blood from the patient. The feature costs are based on the data from Ontario Ministry of Health (1992).

Table 2: Example misclassification cost matrix ($c_{ij}$) for the symmetric and asymmetric misclassification costs. $c_{ij}$ are set in way to achieve a prior expected misclassification cost of 1. In the symmetric cost case, choosing the most probable class leads to $EMC = 1$, whereas, in the asymmetric cost case, the choosing either class is indifferent and both leads to the same $EMC$ of 1.

| Actual Class | Prior Probability | Pred. Class | Symm. Cost | Asymm. Cost |
|---|---|---|---|---|
| $y_1$ | $P(y_1) = 0.6510$ | $y_1$ | 0 | 0 |
|  |  | $y_2$ | 2.866 | 2.866 |
| $y_2$ | $P(y_2) = 0.3490$ | $y_1$ | 2.866 | 1.536 |
|  |  | $y_2$ | 0 | 0 |

We observed that most of the features were assigned the same cost. For example, four out of five features in the Bupa Liver Disorders dataset, 13 out of 19 features in the Hepatitis dataset, six out of eight features in the Diabetes dataset, and 16 out of 20 features in the Thyroid Disease dataset were assigned the same cost. When the costs are so similar, the problem is practically equivalent to finding the minimum size decision tree. To provide more structure into the feature acquisition costs, we also experimented with randomly generated feature and group costs. For each feature, we randomly generated a cost between 1 and 100, and for each group we generated a cost between 100 and 200. We repeated the experiments with three different seeds for each dataset.

The misclassification costs were not defined in the paper by Turney (1995). One reason could be that it is easier to define the feature costs, but defining the cost of a misclassification can be non-trivial. Instead, Turney tests different acquisition strategies using various misclassification costs. We follow a similar technique with a slight modification. We compare the above acquisition policies under both symmetric ($c_{ij} = c_{ji}$) and asymmetric misclassification costs. To be able to judge how the misclassification cost structure affects feature acquisition, we unify the presentation, and compare different acquisition strategies under the same a priori expected misclassification costs, as defined in Equation (1). Specifically, we compare the acquisition policies under various a priori $EMC$ that are achieved by varying the $c_{ij}$ accordingly. We show an example misclassification table for an $EMC$ value of 1 in Table 2. For the real feature cost case, we varied the $EMC$ between 0 and 2000, and varied it from 0 to 4000 for the synthetic feature cost case.

We compare the `Greedy`, `Greedy-LA`, and `Markov Blanket` policies by plotting how much cost each policy saves with respect to the `No Acquisition` policy. In the X axis of the plots, we vary a priori expected misclassification cost using the methodology we described above. We plot the savings on the Y axis. For each dataset, we plot four different scenarios: the cross product of {symmetric, asymmetric} misclassification costs, and {real, synthetic} feature costs.

The results for the Liver Disorders, Diabetes, Heart Disease, Hepatitis, and Thyroid Disease are given in Figures 4, 5, 6, 7, and 8 respectively. For each figure, symmetric misclassification cost scenarios are given in sub-figures (a) and (c), whereas the asymmetric misclassification cost scenarios are presented in (b) and (d). Similarly, the real feature cost scenarios are given in (a) and (b) and the synthetic feature cost scenarios are presented in (c) and (d). We next summarize the results.

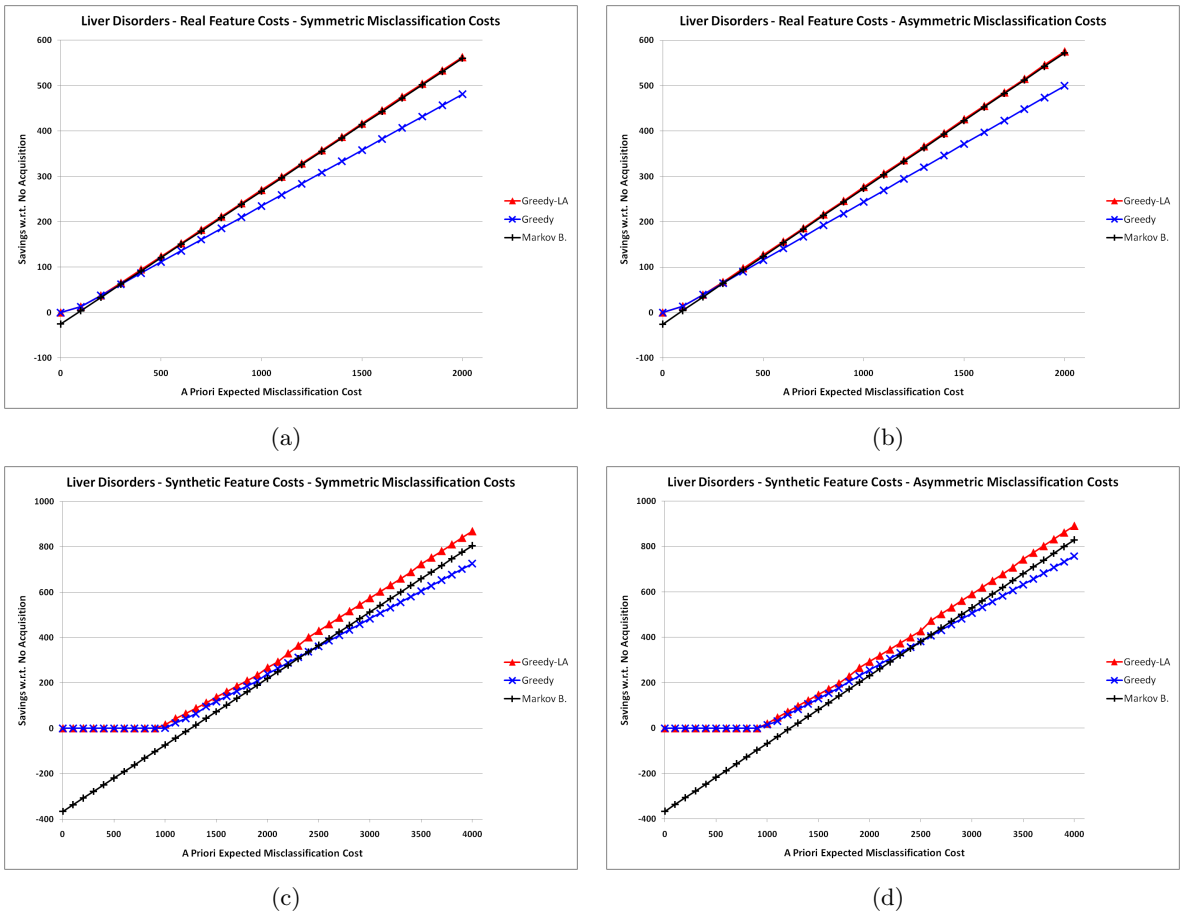

Figure 4: Expected Total Cost ($ETC$) comparisons for the Bupa Liver Disorders dataset. The a priori class distribution is as follows: $P(Y) = [0.4959, 0.5041]$.

- We found that the `Greedy` policy often prematurely stopped acquisition, performing even worse than the `Markov Blanket` strategy. This is true for most of the datasets, regardless of the feature and misclassification cost structures. The fact that the `Greedy` strategy can perform worse than `Markov Blanket` strategy is really troubling. At first, it might seem rather unintuitive that `Greedy` strategy can perform worse than `Markov Blanket` strategy. Part of the reason is that the features belong to groups and the first feature from its group incurs an overhead cost. In `Greedy` strategy where each feature is considered in isolation, the overhead costs can outweigh each single feature's benefit, and because `Greedy` does not look ahead, it is reluctant to commit to acquiring the first feature from any group.

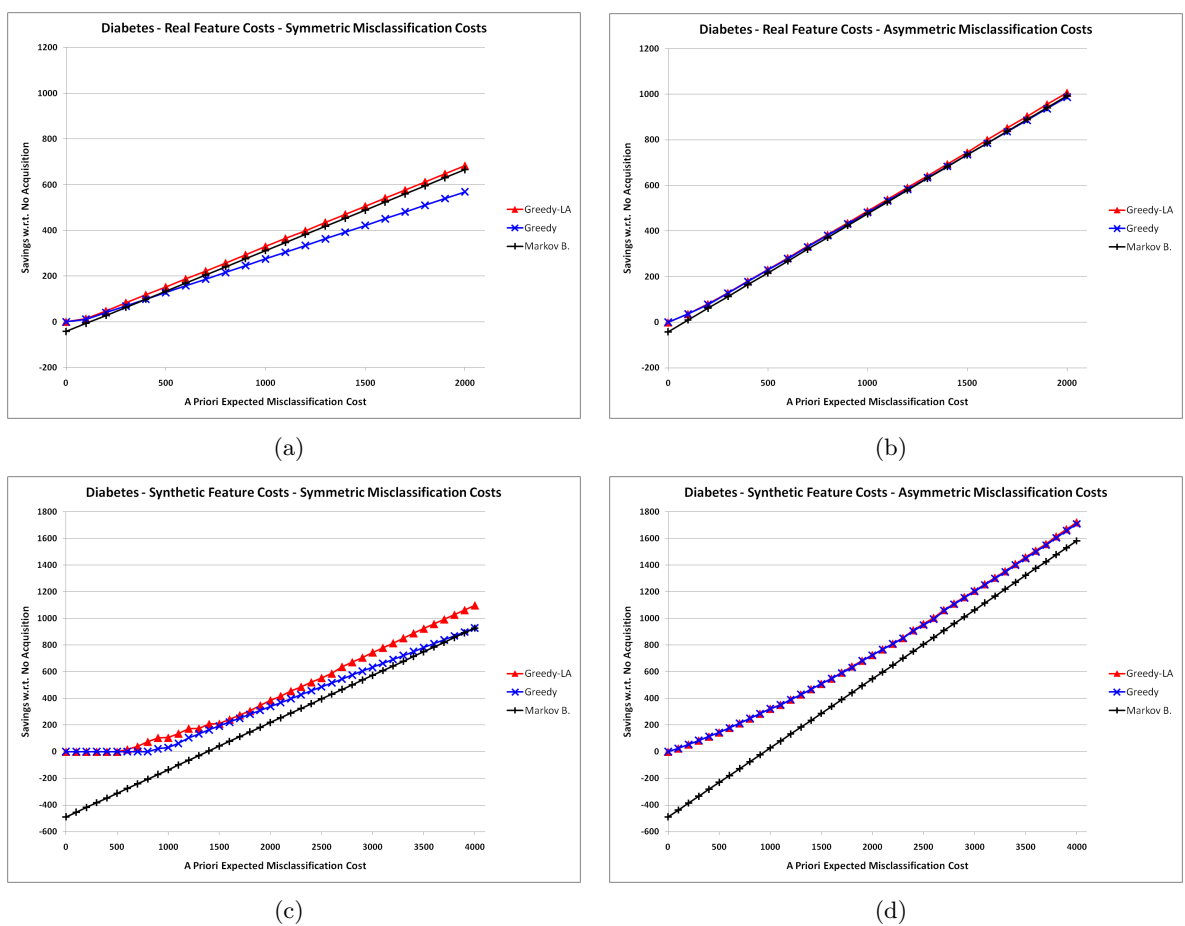

Figure 5: Expected Total Cost ($ETC$) comparisons for the Pima Indian Diabetes dataset. A priori class distribution is as follows: $P(Y) = [0.6510, 0.3490]$.

- `Greedy-LA` strategy never performs worse than any other strategy under any setting.

- The misclassification cost structure (symmetric or asymmetric) had a considerable effect on how the policies behaved. The differences between symmetric and asymmetric cases were particularly evident for datasets where the class distribution was more imbalanced, such as the Diabetes (Figure 5), Hepatitis (Figure 7), and the Thyroid Disease (Figure 8) datasets. The differences due to the misclassification cost structure can be summarized as follows:

    – When the class distribution is imbalanced and the misclassification cost is symmetric, acquiring more information cannot change the classification decisions easily due to the class imbalance, thus the features do not have high $EVI$ values. On the other hand, if the misclassification costs are asymmetric, features tend to have higher $EVI$ values. Thus, the `Greedy` and `Greedy-LA` strategies start acquiring features earlier in the X axis for the asymmetric cases compared to

87

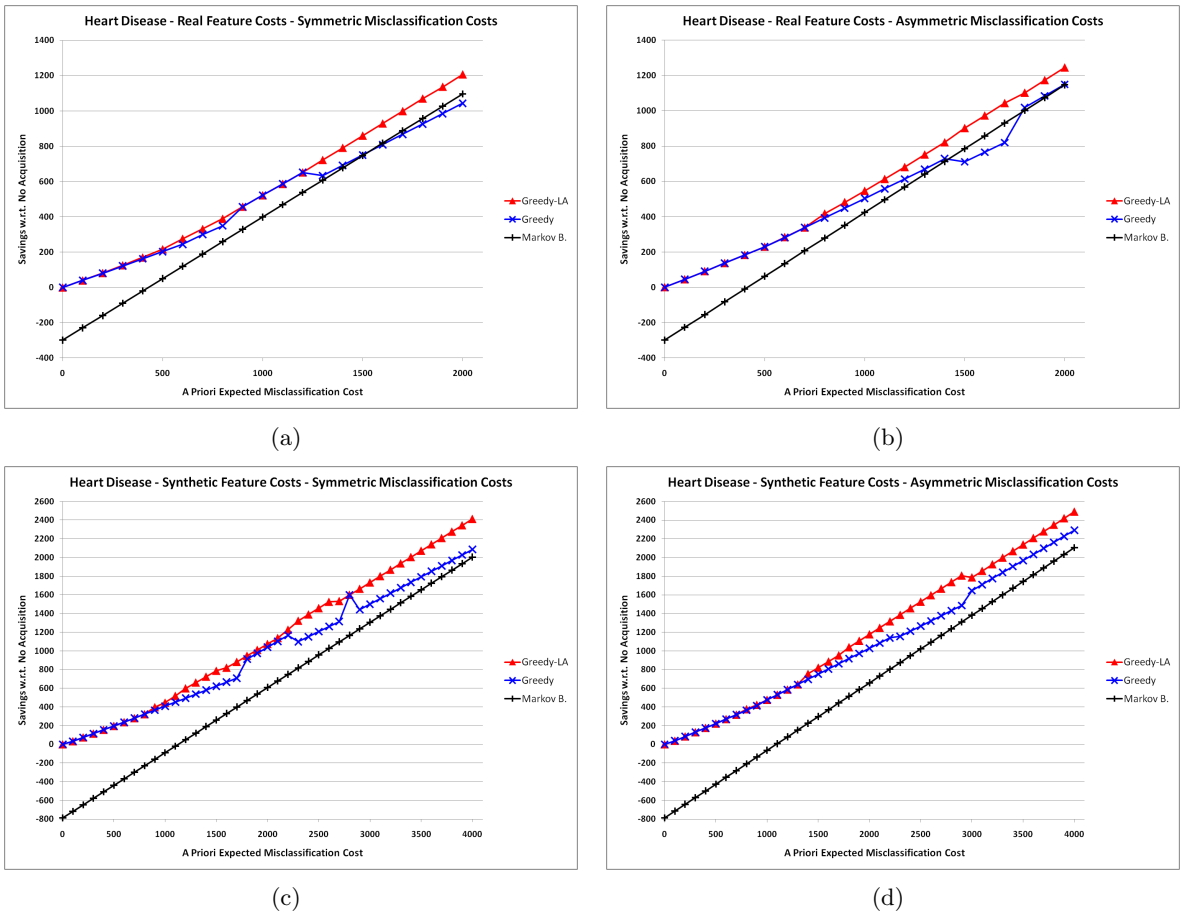

Figure 6: Expected Total Cost ($ETC$) comparisons for the Heart Disease dataset. A priori class distribution is as follows: $P(Y) = [0.5444, 0.4556]$.

their symmetric counterparts. For example, for the Thyroid disease dataset with real feature costs, the `Greedy` strategy starts acquisition only when the $EMC$ is greater than 600 for symmetric misclassification costs (Figure 8(a)) whereas it starts acquiring when the $EMC$ reaches only 100 for the asymmetric case (Figure 8(b)). For the synthetic feature costs, the results are more dramatic; neither `Greedy` or `Greedy-LA` acquires any features for the symmetric cost case (Figure 8(c)), whereas they start acquisition when $EMC = 200$ for the asymmetric case (Figure 8(d)).

– In the same realm with the above results, the slope of the savings for the asymmetric case is much higher compared to the symmetric case.

– The misclassification cost structure causes differences between the `Greedy` and `Greedy-LA` policies in a few cases. For the Diabetes dataset `Greedy` policy performs worse when the misclassification costs are symmetric (Figures 5(a) and 5(c)), whereas for the Hepatitis dataset, it performs worse for the asymmetric misclassification costs (Figures 7(b) and 7(d)).

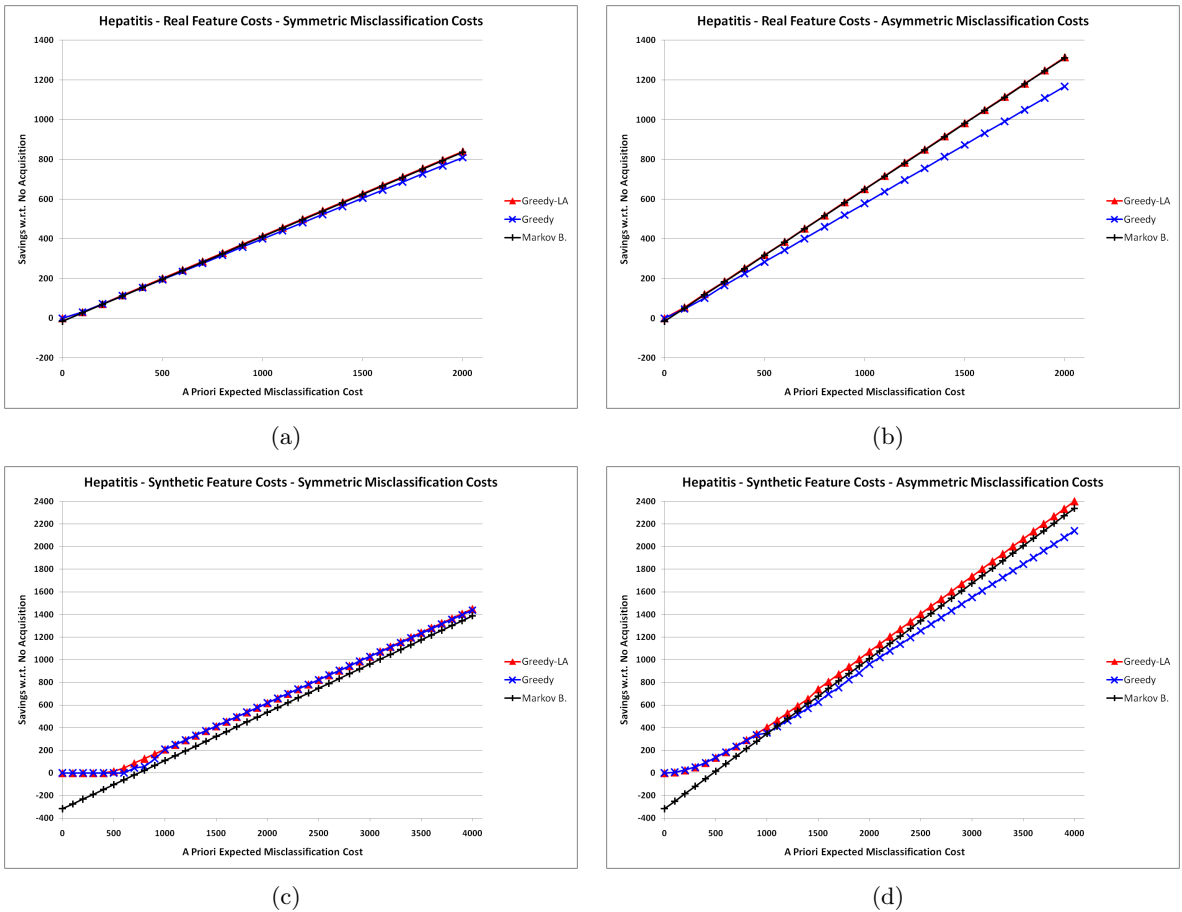

Figure 7: Expected Total Cost ($ETC$) comparisons for the Hepatitis dataset. A priori class distribution is as follows: $P(Y) = [0.7908, 0.2092]$.

- The `Greedy` policy sometimes has an erratic, unpredictable, and unreliable performance as the expected misclassification changes. It possibly hits a local minima, gets out of it later, and hits local minima again (Figures 6 and 8(d)).

We finally present an aggregate summary of the results in Table 3. Table 3 shows how much the `Greedy` policy and the `Greedy-LA` policy saves over the `Markov Blanket` policy. The results are presented as the average saving over various intervals, such as [0-500). As this table also shows, the `Greedy-LA` policy never loses compared to the `Markov Blanket` policy, as one would expect. Additionally, the `Greedy-LA` policy wins over the `Greedy` policy for most of the cases, and it never looses. Finally, `Greedy` policy prematurely stops acquisition, having negative savings with respect to the `Markov Blanket` strategy.

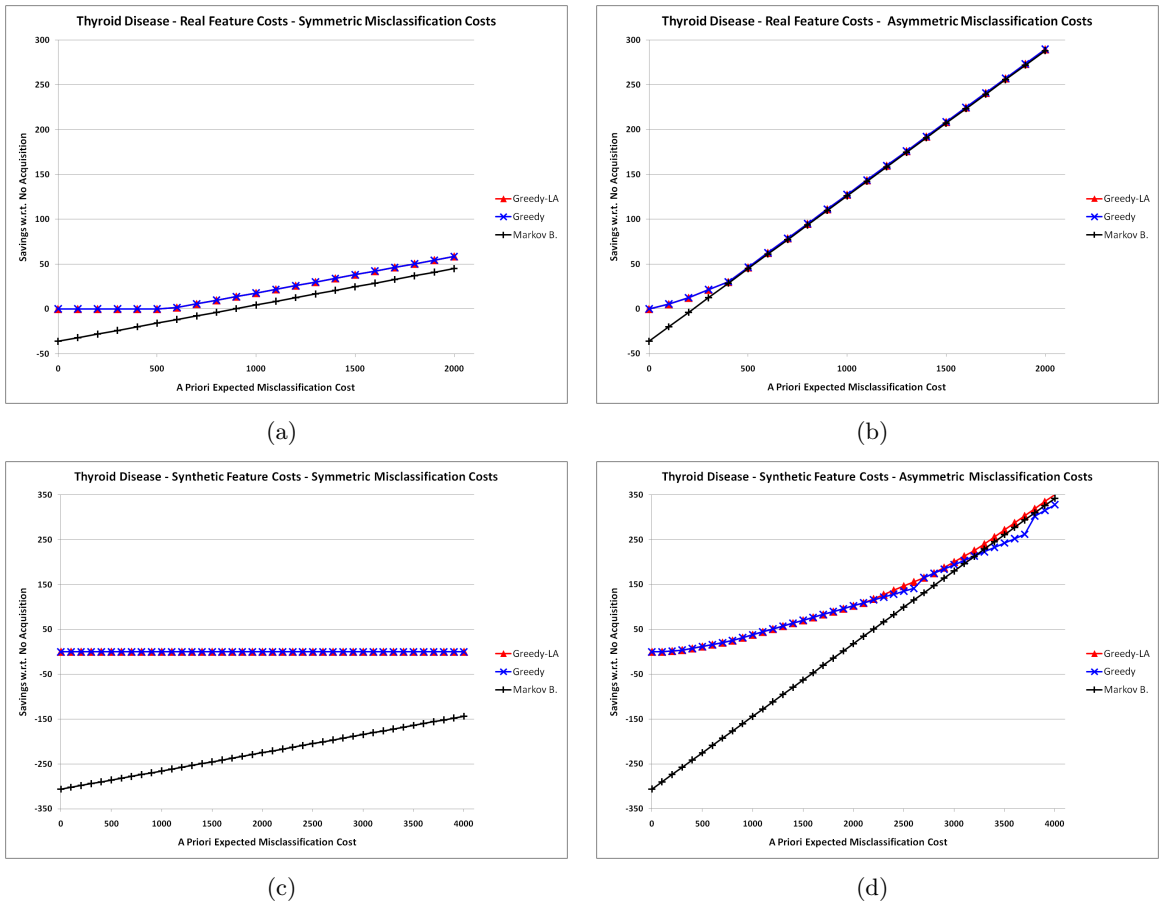

Figure 8: Expected Total Cost ($ETC$) comparisons for the Thyroid Disease dataset. A priori class distribution is as follows: $P(Y) = [0.0244, 0.0507, 0.9249]$.

## 5. Related Work

Decision theoretic value of information calculations provide a principled methodology for information gathering in general (Howard, 1966; Lindley, 1956). Influence diagrams, for example, are popular tools for representing decisions and utility functions (Howard & Matheson, 1984). However, because devising the optimal acquisition policy (i.e., constructing the optimal decision tree) is intractable in general, most of the approaches to feature acquisition have been myopic (Dittmer & Jensen, 1997), greedily acquiring one feature at a time. The greedy approaches typically differ in i) the problem setup they assume, ii) the way the features are scored, and iii) the classification model being learned. We review existing work here, highlighting the differences between different techniques in these three dimensions.

Gaag and Wessels (1993) consider the problem of "evidence" gathering for diagnosis using a Bayesian Network. In their setup, they gather evidence (i.e., observe the values of the variables) until the hypothesis is confirmed or disconfirmed to a desired extent. They propose an acquisition algorithm that greedily computes the expected utility of acquiring a feature and chooses the one with the highest utility. They define the utility as the absolute value of the change in the probability distribution of the hypothesis being tested.

Table 3: Savings of `Greedy` (GR) and `Greedy-LA` (LA) with respect to the `Markov Blanket` policy, averaged over different intervals. An entry is in bold if it is worse than `Greedy-LA`, and it is in red if it is worse than `Markov Blanket`.

|  | Liver | | Diabetes | | Heart | | Hepatitis | | Thyroid | |
|---|---|---|---|---|---|---|---|---|---|---|
|  | GR | LA | GR | LA | GR | LA | GR | LA | GR | LA |
| Real Feature Costs & Symmetric Misclassification Costs | | | | | | | | | | |
| [0-500) | **6.77** | 9.08 | **15.49** | 24.27 | **240.59** | 243.31 | **4.19** | 5.86 | 28.07 | 28.07 |
| [500-1000) | **-18.84** | 2.70 | **-18.28** | 17.06 | **121.31** | 144.87 | **-6.06** | 3.90 | 13.90 | 13.90 |
| [1000-1500) | **-42.12** | 2.66 | **-48.35** | 17.35 | **79.07** | 116.68 | **-14.32** | 3.90 | 13.41 | 13.41 |
| [1500-2000] | **-67.59** | 2.85 | **-81.43** | 17.34 | **-24.98** | 111.34 | **-23.40** | 3.85 | 13.41 | 13.41 |
| Real Feature Costs & Asymmetric Misclassification Costs | | | | | | | | | | |
| [0-500) | **7.33** | 9.3 | **22.74** | 23.84 | 245.79 | 245.79 | **-9.55** | 5.84 | 17.7 | 17.7 |
| [500-1000) | **-16.78** | 2.66 | **9.85** | 13.31 | **131.36** | 143.3 | **-47.61** | 2.57 | 1.56 | 1.56 |
| [1000-1500) | **-38.26** | 3.04 | **3.99** | 11.7 | **46.20** | 114.23 | **-84.79** | 2.57 | 1.56 | 1.56 |
| [1500-2000] | **-61.88** | 2.97 | **-2.54** | 13.7 | **-40.96** | 107.14 | **-125.69** | 2.57 | 1.56 | 1.56 |
| Synthetic Feature Costs & Symmetric Misclassification Costs | | | | | | | | | | |
| [0-500) | 307.39 | 307.39 | 418.34 | 418.34 | 723.36 | 723.36 | 231.93 | 231.93 | 298.01 | 298.01 |
| [500-1000) | 160.95 | 160.95 | **245.75** | 288.65 | **579.25** | 585.72 | **63.59** | 106.54 | 277.70 | 277.70 |
| [1000-1500) | **60.30** | 79.76 | **163.80** | 224.09 | **444.42** | 539.88 | 96.39 | 96.39 | 257.40 | 257.40 |
| [1500-2000) | **31.86** | 53.97 | **138.78** | 163.45 | **378.43** | 490.23 | 88.14 | 88.14 | 237.09 | 237.09 |
| [2000-2500) | **10.43** | 53.01 | **108.69** | 163.78 | **364.03** | 482.24 | 79.88 | 79.88 | 216.79 | 216.79 |
| [2500-3000) | **-14.60** | 62.66 | **78.90** | 164.75 | **268.00** | 458.89 | 71.63 | 71.63 | 196.48 | 196.48 |
| [3000-3500) | **-39.64** | 59.96 | **48.83** | 172.76 | **171.91** | 422.06 | **63.38** | 68.67 | 176.18 | 176.18 |
| [3500-4000] | **-67.18** | 63.68 | **15.75** | 172.13 | **109.91** | 412.11 | **54.30** | 63.66 | 153.84 | 153.84 |
| Synthetic Feature Costs & Asymmetric Misclassification Costs | | | | | | | | | | |
| [0-500) | 306.19 | 306.19 | 441.29 | 441.29 | 728.57 | 728.57 | 219.04 | 219.04 | 276.32 | 276.32 |
| [500-1000) | 156.78 | 156.78 | 341.28 | 341.28 | **599.02** | 603.12 | **88.34** | 91.53 | 213.60 | 213.60 |
| [1000-1500) | **66.57** | 79.79 | **260.09** | 261.21 | **505.80** | 517.37 | **-16.86** | 49.39 | 162.52 | 162.52 |
| [1500-2000) | **37.47** | 60.62 | **201.19** | 204.31 | **420.56** | 519.29 | **-54.31** | 61.90 | 113.82 | 113.82 |
| [2000-2500) | **14.84** | 55.70 | **161.24** | 164.17 | **320.32** | 512.75 | **-64.75** | 61.90 | **65.12** | 68.39 |
| [2500-3000) | **-9.19** | 58.85 | **144.24** | 151.22 | **211.26** | 500.75 | **-101.93** | 61.90 | **28.72** | 34.73 |
| [3000-3500) | **-33.22** | 59.33 | **132.84** | 139.54 | **248.73** | 400.16 | **-139.11** | 61.90 | **0.78** | 14.67 |
| [3500-4000] | **-59.66** | 63.13 | **126.43** | 136.51 | **206.06** | 389.32 | **-180.01** | 61.90 | **-18.10** | 9.50 |

In more recent work, Sent and Gaag (2007) consider the problem of acquiring more than a single feature at each step. They define subgoals and cluster the features for each subgoal. The subgoals and clustering of the features are provided by the domain experts. Then, they in the non-myopic case, they pick a cluster by calculating their expected values. However, because clusters can be big, calculating the expected value of a cluster can be problematic; thus, they also provide a semi-myopic algorithm where they pick the cluster that has the best (myopic) feature.

Núñez (1991) introduces a decision tree algorithm called EG2 that is sensitive to the feature costs. Rather than splitting the decision tree at a feature that has high information gain, EG2 chooses a feature that has least "information cost function," which is defined as the ratio of a feature's cost to its discriminative efficiency. EG2 is, however, is not directly optimized to balance the misclassification cost and feature acquisition cost; rather it is optimized for 0/1 loss while taking the feature costs into account. Similarly, Tan (1990) modifies the ID3 algorithm (Quinlan, 1986) to account for feature costs. Tan considers the domain where a robot needs to sense, recognize, and act, and the number of features is very large. For the robot to act efficiently, it needs to trade-off accuracy for efficiency.

Turney (1995) builds a decision tree called ICET (standing for Inexpensive Classification with Expensive Tests) using a genetic search algorithm (Grefenstette, 1986) and using Núñez's (1991) criteria to build C4.5 decision trees (Quinlan, 1993). Unlike Núñez, Turney takes misclassification costs into account (in addition to the feature costs) to evaluate a given decision tree and looks for a good decision tree using genetic search algorithms.

Yang et al. (2006) build cost-sensitive decision trees and Naive Bayes classifiers that take both feature costs and misclassification costs into account. Unlike Núñez (1991), who scores features based on information gain and cost ratio, Yang et al. score features based on expected reduction in the total cost (i.e., sum of the feature cost and the misclassification cost) on the training data. By doing so, they take feature costs and misclassification costs into account directly at learning time.

Bayer-Zubek (2004) formulates the feature acquisition problem as a Markov Decision Process and provides both greedy and systematic search algorithms to develop diagnostic policies. Bayer-Zubek takes both feature cost and misclassification costs into account and automatically finds an acquisition plan that balances the two costs. She introduces an admissible heuristic for AO* search and describes regularization techniques to reduce overfitting to the training data.

Saar-Tsechansky, Melville, and Provost (2009) consider active feature acquisition for classifier induction. Specifically, they are given a training data with missing feature values, and a cost matrix that defines the cost of acquiring each feature value, they describe an incremental algorithm that can select the best feature to acquire iteratively to build a model that is expected to have high future performance. The utility of acquiring a feature is estimated in terms of expected performance improvement per unit cost. The two characteristics that make this work different from most of the previous work is that i) the authors do not assume a fixed budget a priori; rather they build the model incrementally, ii) each feature can have a different cost for each instance.

Finally, Greiner, Grove, and Roth (2002) analyze the sample complexity of dynamic programming algorithms that performs value iteration to search for the best diagnostic policies. They analyze the problem of learning the optimal policy, using a variant of the probably-approximately-correct (PAC) model. They show that the learning can be achieved efficiently when the active classifier is allowed to perform only (at most) a constant number of tests and show that learning the optimal policy is often intractable in more general environments.

## 6. Future Work

In this article, we have only scratched the surface of incorporating constraints between features in order to reduce the search space and make reasoning with sets tractable. We have discovered two types of constraints (features that render other features useless, and features that are useless without other features) purely from the underlying probability distribution. We have shown that these automatically discovered constraints helped reduce the search space dramatically. In practice, it is possible to discover additional types of constraints that can potentially be used reduce the search space further (for e.g., ordering constraints where certain procedures always precede other procedures). Constraints can also be defined based on observed feature values; for example, a treadmill test might not be performed for patients of old age. Patients can decline certain procedures and medications. Eliciting these constraints from the domain experts and utilizing them to further reduce the search space is a promising future direction.

Most of the existing feature acquisition frameworks, including this one, are a major simplification of what happens in practice; we have assumed that acquiring the values of the features does not change the class value or values of other variables. However, in practice, feature value measurements can have "side-effects," for example, in medical diagnosis while certain measurements are non-invasive and do not change the status of the patient, others might include medications that can affect the outcome. Similarly, in fault diagnosis and repair, the purpose is not only to diagnose but it is to repair the fault, so some actions can in fact repair the fault, in essence changing the class value. Taking these extra "side-effects" into account will make feature acquisition frameworks more realistic.

## 7. Conclusion

The typical approach to feature acquisition has been greedy in the past primarily due to the sheer size of the possible subsets of features. We described a general technique that can optimally prune the search space by exploiting the conditional independence relationships between the features and the class variable. We empirically showed that exploiting the conditional independence relationships can substantially reduce the number of possible subsets. We also introduced a novel data structure called Value of Information Lattice (`VOILA`) that can both efficiently reduce the search space using the conditional independence relationships and also can share probabilistic inference computations between different subsets of features. By using `VOILA`, we were able to add a full look-ahead capability to the greedy acquisition policy, which would not be practical otherwise. We experimentally showed on five real-world medical datasets that the greedy strategy often stopped feature acquisition prematurely, performing worse than even a policy that acquires all the features.

## Acknowledgments

We thank the reviewers for their helpful and constructive feedback. This material is based on work supported by the National Science Foundation under Grant No. 0746930.